# A Deep Reinforcement Learning Approach for Ramp Metering Based on Traffic Video Data

Bing Liu,[1] Yu Tang,[2] Yuxiong Ji,[1] Yu Shen,[1] and Yuchuan Du[1]

[1] Key Laboratory of Road and Traffic Engineering of the Ministry of Education, Tongji University, Shanghai 201804, China.
[2] Tandon School of Engineering, New York University, New York 11201, USA.

Correspondence should be addressed to Yu Shen; yshen@tongji.edu.cn

## Abstract

Ramp metering that uses traffic signals to regulate vehicle flows from the on-ramps has been widely implemented to improve vehicle mobility of the freeway. Previous studies generally update signal timings in real-time based on predefined traffic measures collected by point detectors, such as traffic volumes and occupancies. Comparing with point detectors, traffic cameras—which have been increasingly deployed on road networks—could cover larger areas and provide more detailed traffic information. In this work, we propose a deep reinforcement learning (DRL) method to explore the potential of traffic video data in improving the efficiency of ramp metering. The proposed method uses traffic video frames as inputs and learns the optimal control strategies directly from the high-dimensional visual inputs. A real-world case study demonstrates that, in comparison with a state-of-the-practice method, the proposed DRL method results in 1) lower travel times in the mainline, 2) shorter vehicle queues at the on-ramp, and 3) higher traffic flows downstream of the merging area. The results suggest that the proposed method is able to extract useful information from the video data for better ramp metering controls.

*Keywords*: Ramp metering, Deep *Q*-learning, Traffic videos

# Introduction

Ramp metering uses traffic signals to regulate vehicle flows from on-ramps to the mainline of the freeway. It alleviates the negative impacts of "capacity drop" resulting from massive merging behaviors and reduces the total time spent in the traffic system [1, 2]. Several field tests have demonstrated the effectiveness of ramp metering in terms of throughput, vehicle-miles-traveled, vehicle-hours-traveled, and travel time reliability [3-6]. For instance, a large-scale experiment conducted in Minneapolis-Saint Paul found that when the ramp meters were turned off, freeway capacity decreased by 9%, travel time increased by 22%, and crashes increased by 26% [7].

Most of the ramp metering methods are driven by traffic measures, such as volumes, queue lengths, and occupancies, obtained from point detectors (*e.g.*, inductive loop detectors) [8, 9]. These measures are predefined and only reflect partial information regarding traffic operations. Recently, traffic cameras have been increasingly deployed on road networks for monitoring traffic operations and detecting illegal driving behaviors. Compared with point detectors, traffic cameras cover larger areas and provide more detailed traffic information, such as vehicle locations, vehicle speeds, and headways between vehicles.

We propose a deep reinforcement learning (DRL) method to explore the potential of traffic video data in improving the efficiency of ramp metering. The proposed method does not rely on hand-crafted traffic measures. Instead, it uses traffic video frames as inputs and learns the optimal control strategies directly from high-dimensional visual inputs by taking advantage of the special neutral structures in deep learning, which are capable of automatically extracting high-level features from raw data. The effectiveness of the proposed method is demonstrated in comparison with a state-of-the-practice method in a real-world case study.

This paper is organized as follows. A brief review of related literature is presented first, and then the methodology is described, followed by the demonstration in a real-world case study. The final section summarizes the paper and discusses possible directions for future research.

# Related Works

The signal timing plans for ramp metering could be fixed or traffic responsive. Wattleworth [10] firstly formulated a linear programming model to optimize a fixed signal timing plan for ramp metering based on historical traffic volume data. Responsive approaches adapt to traffic flow fluctuations by updating signal timings in response to real-time traffic measures. The responsive approaches can be classified as rule-based, optimization-based, and RL-based.

The rule-based approaches update signal timings in real-time according to specific rules. Masher et al. [11] proposed a feedforward ramp metering algorithm, aiming at keeping the downstream traffic flow below the capacity. Papageorgiou et al. [8] proposed a feedback ramp metering algorithm, which is named as ALINEA, based on the occupancy obtained by inductive loop detectors downstream of the merging area [12]. The ALINEA regulates the inflows from the on-ramp in the scheme of closed-loop control so that the detected occupancy would approach a predefined target value. The ALINEA was further extended by considering more traffic measures, such as queue lengths at the on-ramp and traffic volumes in the mainline [13]. Wang et al. [9] proposed the proportional-integral-based ALINEA (i.e. PI-ALINEA), considering the cases where a bottleneck with a capacity lower than that of the merging area is present further downstream of the mearing area. They demonstrated numerically that the PI-ALINEA performs better than the original ALINEA. The ALINEA has been incorporated in several coordinated ramp metering algorithms, such as METALINE and HERO [14, 15].

The optimization-based approaches optimize signal timings based on real-time traffic data. The model predictive control framework, which considers the interactions between ramp metering and future traffic states, is often employed to predict traffic evolution for proactive traffic controls. Nevertheless, the models used to describe traffic dynamics may lead to nonlinear control problems, which brings the difficulty in finding the optimal signal timings [16-18]. In addition, the effectiveness of the ramp metering strategy depends on the degree of the fitness of the models to the actual traffic dynamics.

The RL based approaches search for a policy that determines the signal timings based on the current traffic state. Existing RL-based ramp metering approaches were mainly developed based on value-based RL methods, such as the Q-learning algorithm. They evaluate the value of policy given each state-action pair and improve the policy with a heuristic search. The actions usually refer to the metering rate or signal phase selection. The states are represented by traffic measures, such as traffic density and volume. The downstream and upstream traffic flow, ramp queue length, and traffic density are often selected to define the control reward [19-22]. The RL has also been introduced for coordinated ramp metering [23, 24]. Note that existing RL methods for ramp metering were driven by traditional traffic measures. To the best of our knowledge, no studies have attempted to automatically extract information from traffic videos and learns the optimal control strategies directly from the visual inputs for ramp metering. Some studies have considered traffic video data as inputs in the RL framework for intersection signal controls [25-27]. Nevertheless, the elements in the RL framework, such as the action, reward, and state, for ramp metering and intersection signal controls are different.

## Methodology

We propose a DRL method for local ramp metering based on traffic video data. The proposed method adopts a flexible two-phase control scheme, which is illustrated in Fig. 1. A policy trained from the DRL determines whether the phase in the next time step is green or red at a fixed time step $L$ based on current traffic state. The adopted control scheme is conceivably more flexible than the conventional "stop-and-go" scheme with fixed red and green phases flashing alternately. To reduce the computation burden, traffic video frames are used as control inputs. Vehicle locations from raw images are extracted to reconstruct visual representations in the DRL method [25].

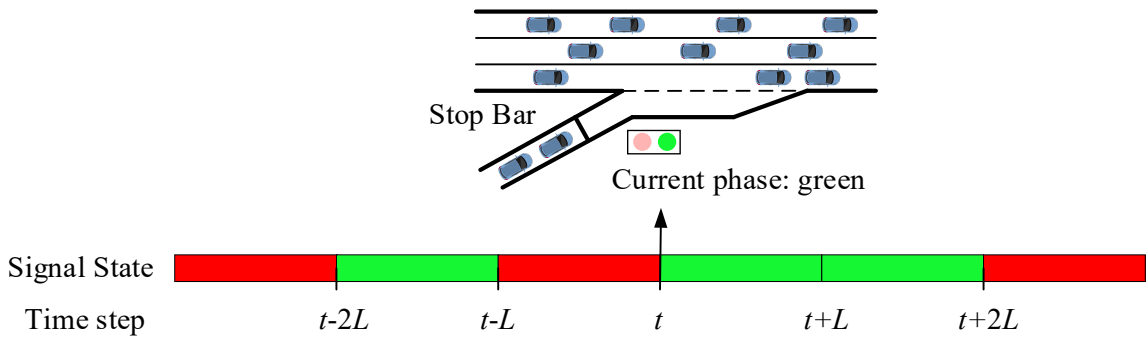

Figure 1: Two-phase control scheme for ramp metering.

Fig. 2 illustrates the ramp metering problem in the general scheme of RL. The ramp meter acts as the agent to interact with the environment, namely the traffic system including the mainline and the on-ramp. The ramp meter takes certain control action $a$ based on current traffic state *s*. Then the traffic system responds to the control action with the state transition from *s* to *s'*. The ramp meter obtains reward *r* that quantifies the effect of the control action *a* given state *s*. In the RL field, the interaction process is assumed to be a Markov Decision Process [28]. That is, given current state *s* and action *a*, the following state *s'* and reward *r* are independent of previous states and actions.

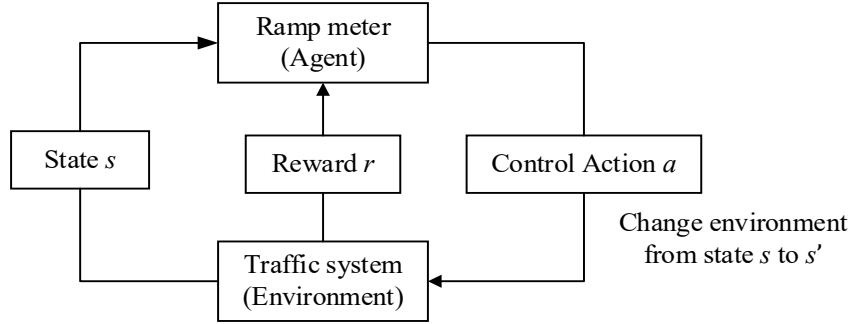

Figure 2: Ramp metering in the general scheme of RL.

The RL method aims at training the agent (Ramp meter) to find an optimal policy $\pi^*$ that maximizes the discounted cumulative reward $G_t$ after time step *t*. To achieve that goal, one approach is to learn the optimal action-value function $Q^*(s, a)$, which is defined by:

$$Q^*(s,a) = \max_{\pi} \mathbb{E}_{\pi}\left[G_t \mid s_t = s, a_t = a\right] \quad (1)$$

where $\pi$ is a policy mapping traffic states to control actions. Once $Q^*(s, a)$ is found, the action to be taken can be obtained by:

$$\pi^*(s) = \mathrm{argmax}_{a'} Q^*(s, a') \quad (2)$$

The process of learning policy from the action-value function $Q$ is termed *Q*-learning. The key step in *Q*-learning is to estimate $Q^*(s, a)$. To better capture the traffic features in video data, we adopt a deep neural network termed *Q*-network to approximate $Q^*(s, a)$. The neural network contains serval convolutional layers and fully-connected layers. The convolutional layers extract traffic features from the video frames and the full-connected layers estimate the Q value based on the output of convolutional layers.

**State, Action, and Reward Representation**

1) Action

The ramp meter takes two possible actions at fixed time step *L* to determine whether the phase in the next time step is green (G) or red (R), namely action set $\mathcal{A} = \{G, R\}$.

2) State

Fig. 3 illustrates the process of obtaining visual representations from the raw images to represent traffic states. Considering a control area of size $X \times Y$ as shown in Fig. 3a, the position of each vehicle *i* $(x_i, y_i)$ is extracted from the raw images. Traffic state at times step *t* is represented by a position matrix $m_t \in \mathbb{R}^{X \times Y}$:

$$m_t(x_i, y_i) = U_V, i \in I_t \quad (3)$$

$U_V$ is a constant and $I_t$ refers to the vehicles set.

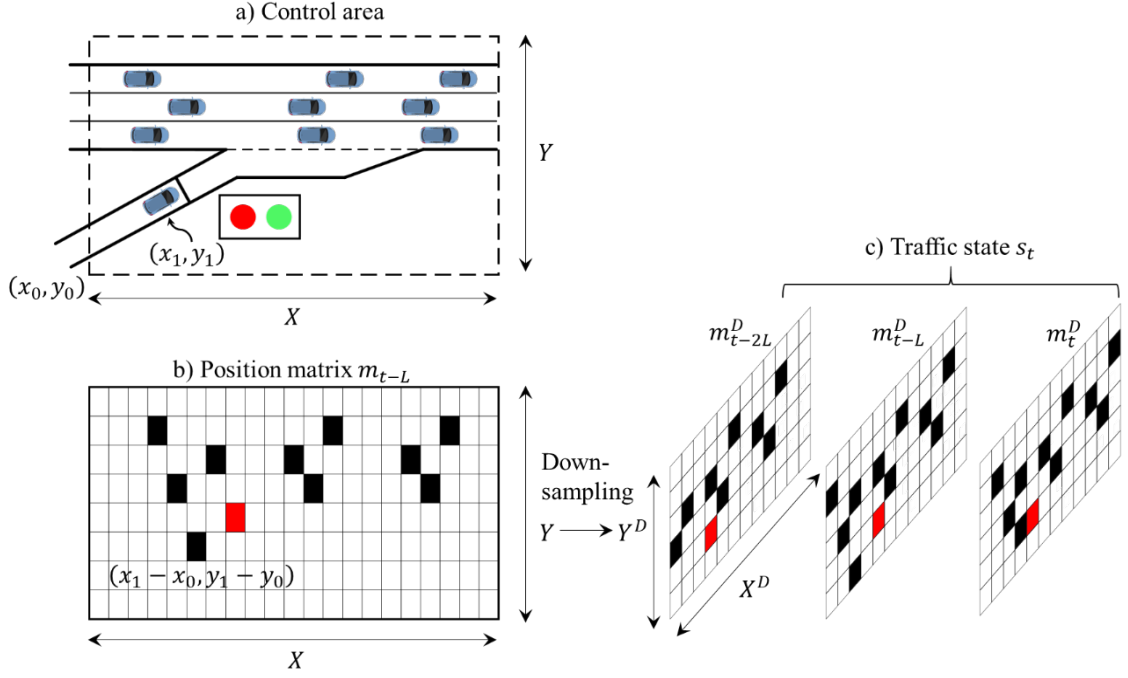

Figure 3: State representation.

The position matrix additionally contains the information of the signal states:

$$m_t(x_R, y_R) = \begin{cases} U_G \text{ if } a_{t-1} = G \\ U_R \text{ if } a_{t-1} = R \end{cases} \quad (4)$$

$(x_R, y_R)$ is the position of signal light in images and $U_G$ and $U_R$ are constants to represent actions execute in time step $t$-1.

The other elements in matrix $m_t$ are set to zero:

$$m_t(x, y) = 0, (x, y) \notin \{(x_i, y_i) | i \in I_t\} \cup \{(x_R, y_R)\} \quad (5)$$

Since one matrix is insufficient to describe vehicle dynamics, we stack the matrices of consecutive $N$ time steps to represent the state $s_t$:

$$s_t = \{m_{t-L(N-1)}, \cdots, m_t\} \quad (6)$$

For computation efficiency, down-sampling is introduced before the matrix stack to reduce the input dimensions and maintain sufficient information about vehicle positions.

3) Reward

The reward is critical since it motivates the agent to approach the objective. Vehicle travel times through the system are good measures of vehicle mobility. But they are not suitable rewards since they are greatly influenced by previous actions and cannot well quantify the value of the current action. Alternatively, the speed in the merging area and the queue length at the on-ramp are adopted to define the reward. The reward $r_t$ after action $a_t$ is defined by:

$$r_t = \sum_{j=t+1}^{t+L} (\mu v_j + \omega q_j)/L, \mu > 0, \omega < 0 \quad (7)$$

where $v_j$ and $q_j$ represent the speed in the merging area and the queue length at the on-ramp at time step $j$, respectively. Positive $\mu$ and negative $\omega$ denote the reward weights for the speed and the queue length, respectively. The defined rewards balance the priority needs between improving vehicle mobility in the mainline and reducing vehicle delays at the on-ramp.

## Algorithm

The training problem can be viewed as the regression problem with the loss function defined by:

$$L(w_i) = \mathbb{E}_{s,a}\left[(Q^*(s,a) - Q(s,a,w_i))^2\right] \quad (8)$$

where $Q^*(s, a)$ is approximated by a deep neural network. $w_i$ denotes the weights in the deep neural network after the $i^{th}$ update. The gradient-based algorithm ADAM [29] is used to update $w_i$ to minimize the loss function.

Although Q-network could handle large state space, it raises the problem of instability in training [30]. Two approaches have been proposed to resolve the problem, illustrated in Fig. 4. The experience replay [31] stores the data prepared for training into the replay buffer and samples these data uniformly for training the agent. It contributes to making the training data independently and identically distributed. The target network [30] is introduced for a stable estimation of target value $Q^*(s, a)$. When Q-network is being trained, the target network is freezing. Its parameters are updated by copying those in Q-network with a certain frequency.

A multi-task learning strategy is implemented in the training process to improve learning efficiency. While training the ramp metering Q-network, the mean speed in the merging area and the queue length at the on-ramp are predicted simultaneously. The fully-connected layer in the Q-network in multi-task learning is extended to two layers to calculate the Q-value and predict the speed and queue length, respectively. The loss function is redefined as follows.

$$\mathcal{L}_1(w_i) = \mathbb{E}_{s,a}(Q^*(s,a) - Q^0(s,a,w_i))^2 \quad (8)$$
$$\mathcal{L}_2(w_i) = \mathbb{E}_{s,a}((Q^1(s,a,w_i) - v)^2 + (Q^2(s,a,w_i) - u)^2) \quad (9)$$
$$\mathcal{L}(w_i) = \mathcal{L}_1(w_i) + \lambda \mathcal{L}_2(w_i) \quad (10)$$

where $Q^0(s,a,w_i)$, $Q^1(s,a,w_i)$, $Q^2(s,a,w_i)$ represent the prediction values of the action-value function, mean speed, and queue length, respectively. $u$ and $v$ refer to the ground truth of the speed and queue length, respectively.

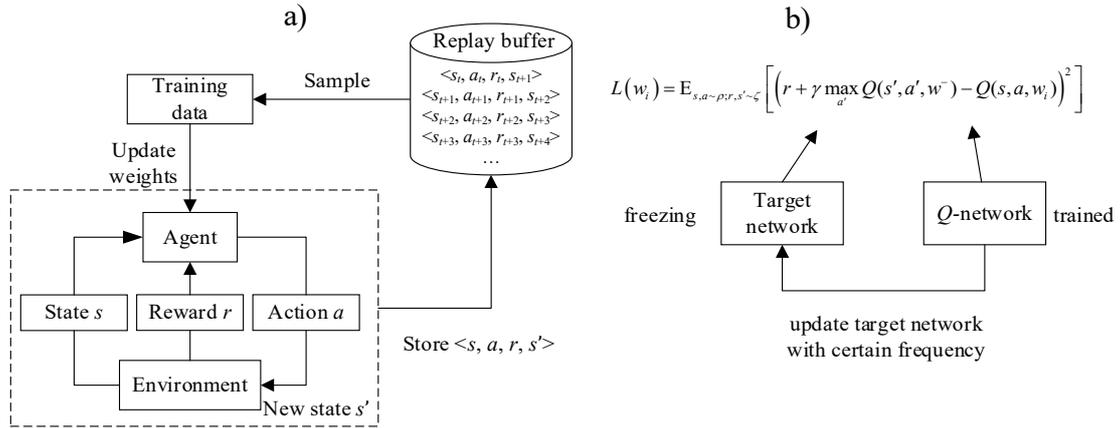

Figure 4: 4 Approaches for stabilizing the training process in deep Q-learning: a) relay buffer; b) target network.

The algorithm for training the ramp metering policy is presented in Table 1. The ramp meter takes new action with the ε-greedy strategy, meaning that the agent does not always adopt the best action derived from current $Q(s, a, w)$. Doing so is helpful to explore unknown policies that may be better.

Table 1: Deep Q-learning for ramp metering.

---

**Input:** Batch size $k$, Learning rate $\eta$, Decision step size $L$, Exploration rate $\varepsilon$,
      Freezing interval $F$, Number of training episode $N$,
      Number of training time steps $T$ in one episode
Fill the replay buffer randomly
Initialize the action-value function $Q$ with random weights $w$
Initialize the target network $Q^-$ with weights $w^- = w$
$f \leftarrow 0$
**for** episode = $\{1, \cdots, N\}$
    **for** $t = \{1, \cdots, T\}$
      **if** mod $(t, L) == 0$
        Observe state $s_t$ and choose action $a_t$ by $\varepsilon$-greedy strategy
        Execute action $a_t$, obtain reward $r_t$, state $s_{t+L}$
        Store transition $<s_t, a_t, r_t, v_t, u_t, s_{t+L}>$ into replay buffer
        Uniformly sample $k$ transitions from replay buffer
        Compute the gradient $\nabla_w(\mathcal{L}_1 + \lambda\mathcal{L}_2)$
        Update weights $w$ in $Q$-network by ADAM with learning rate $\eta$
        $f \leftarrow f + 1$
      **end if**
      **if** mod $(f, F) == 0$
        Update target network $Q^-$ : $w^- \leftarrow w$
      **end if**
    **end for**
**end for**

---

## Case study

**Case set-up**

The proposed method is evaluated in SUMO, an open-source microscopic traffic simulator [32]. The simulation is performed in the freeway connecting Qingdao and Huangdao through a tunnel in Shandong, China, as shown in Fig. 5. The one-lane on-ramp and the upstream three-lane mainline merge. And the freeway gradually reduces to three lanes in the downstream segments.

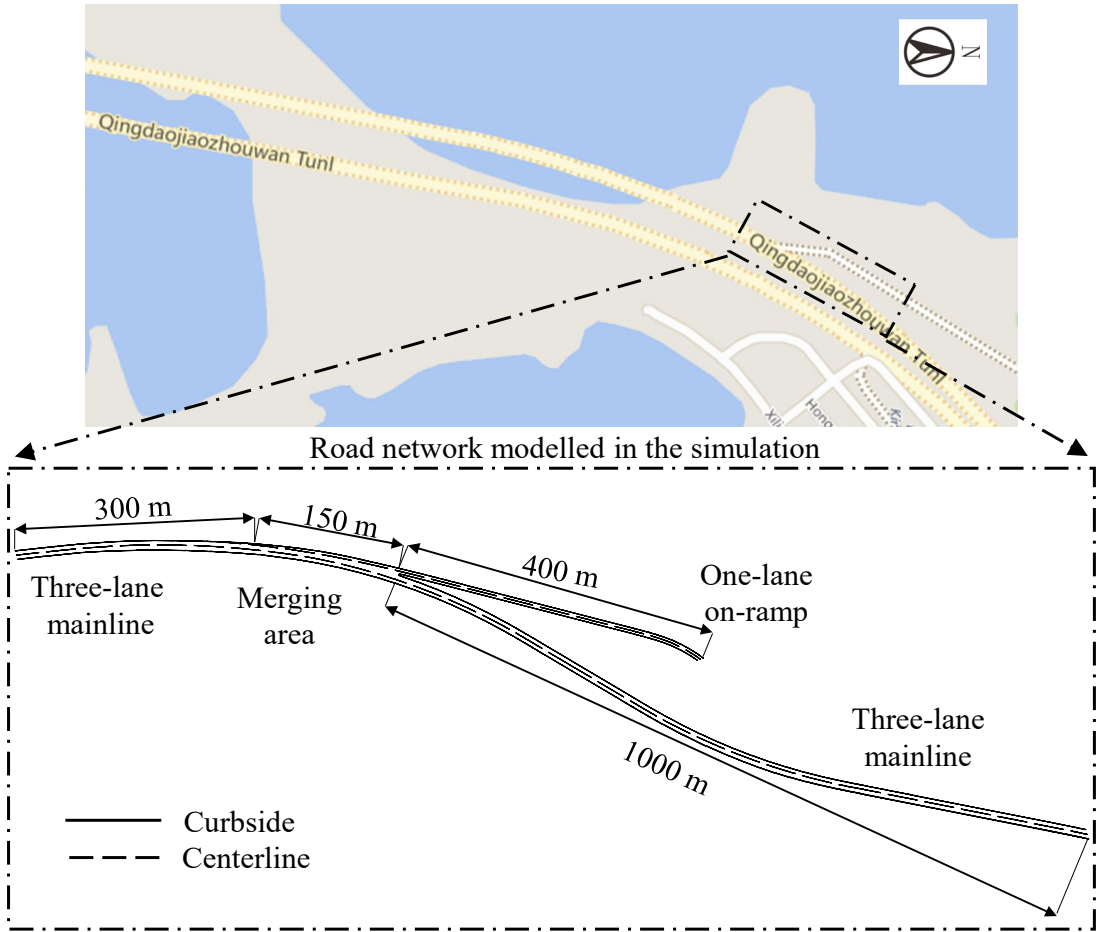

Figure 5: Road network considered in the case study.

The simulation considers the road network covering the on-ramp, the upstream and downstream mainlines, and the merging area. The speed limits in the mainline and at the on-ramp are 80 and 40 km/h, respectively. The simulation parameters regarding driving behaviors are calibrated based on empirical data. The simulation time is 7:50 a.m. to 9:00 a.m. Figure 6 presents the 10-min traffic volumes from 7:50 a.m. to 9:00 a.m. in the mainline and at the on-ramp.

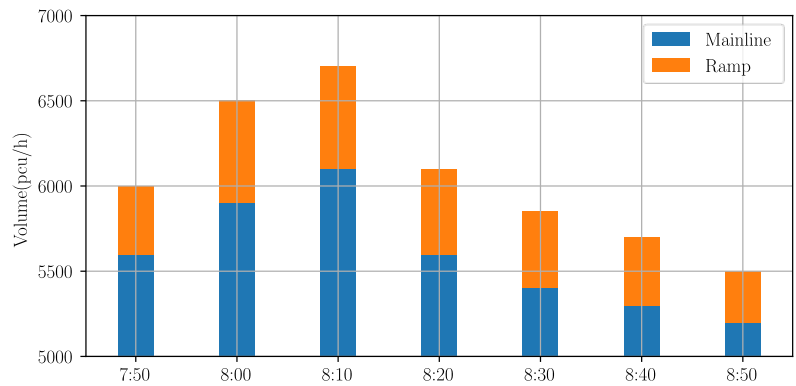

Figure 6: Traffic volumes in the mainline and the on-ramp.

The agent's observation in DRL covers the simulated road network. After zooming and down-sampling, the size of one observation is 4×512 in pixels. One observation is stacked with those in the previous two steps to represent the traffic state. The $Q$-network in [31] is revised as Fig. 7 to accommodate the input size for ramp metering. The other parameters in the deep $Q$-learning for ramp metering are listed in Table 2.

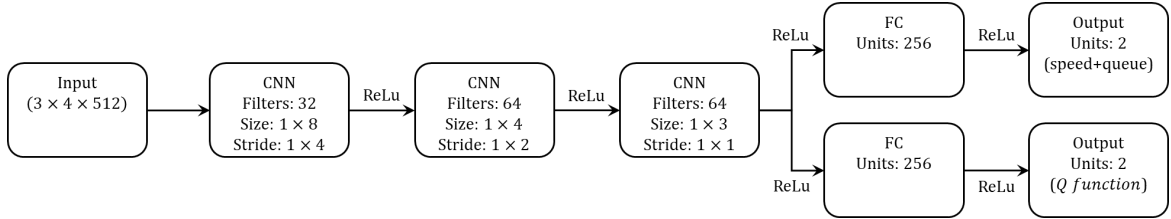

Figure 7: Structure of the deep neural network.

Table 2: Key parameters in Deep $Q$-learning for Ramp Metering

| Parameters | Value |
| --- | --- |
| Replay buffer size $B$ | $2\times10^5$ |
| Learning rate $\eta$ | $2.5\times10^{-4}$ |
| Batch size $k$ | 32 |
| Exploration rate $\varepsilon$ | 0.1 |
| Freeze interval $F$ | $10^4$ |
| Decision step size $L$ | 4 s |
| Reward weight $\mu$ | 0.5 |
| Reward weight $\omega$ | -0.1 |

The $Q$-network in the proposed method is trained in $10^6$ training frames. For comparison purposes, two scenarios are also considered in the evaluation. The first scenario does not apply ramp metering and the second scenario adopts the PI-ALINEA method. The PI-ALINEA method fixes the length of the green phase and determines the length of the red phase in the current signal cycle based on the length of the red phase in the previous cycle and the downstream occupancies in the previous and current cycles. The parameters are fine-tuned to minimize mean travel times in the traffic system. Each scenario is evaluated in 20 simulation experiments, each of which is initialized with different random seeds.

**Evaluation and Comparison**

Fig. 8 presents the Q1 (25%), Q2 (50%), and Q3 (75%) of the travel times in the mainline for three scenarios. The results demonstrate that ramp metering not only reduces vehicle travel times but also improves the stability of traffic flows in the mainline. The resulting travel times are close when the demand is low at the beginning (8:00-8:10). With the increase of the demand, the median travel time increases faster in the no-control scenario and reaches the maximum value of 4.25 min at 8:28. The maximum values of the median travel times resulting from the PI-ALINEA method and the DRL method are 3.75 min and 3.4 min, respectively, which are 11.7% and 20% lower than that in the no-control scenario. As the demand decreases, the travel times of the three scenarios decrease to similar levels. In addition, all the ramp metering methods narrow the interquartile ranges (Q3-Q1) of the travel times. Overall, the proposed DRL method results in shorter travel times with narrower travel time ranges than that of the fine-tuned PI-ALINEA method.

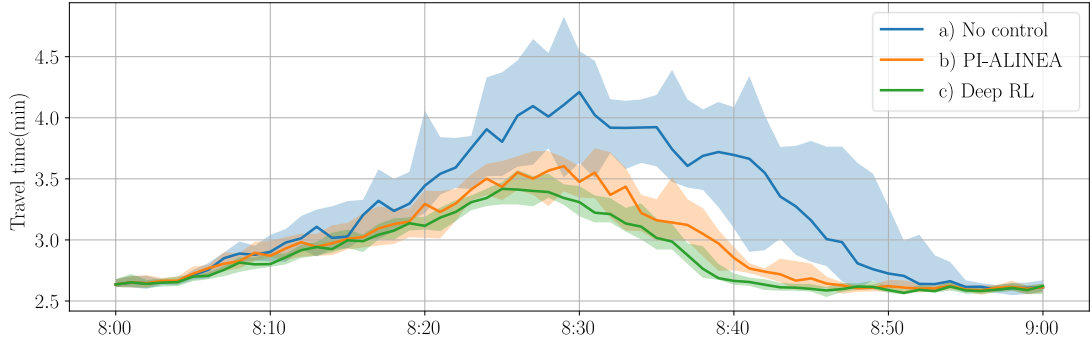

Figure 8: Vehicle travel times in the mainline.

Table 3 presents mean vehicle travel times in 20 simulation experiments. It is found that the performance of the PI-ALINEA method is not always better than the no-control scenario. For example, compared with the no-control scenario, the PI-ALINEA method increases vehicle travel time by 10% in the 6$^{th}$ experiment. In contrast, the proposed DRL method consistently results in shorter travel time than that of the no-control scenario, demonstrating the stable performance of the proposed method. On average, the DRL method reduces vehicle travel time by 13% when compared with the no-control scenario.

Table 3: Mean travel times in 20 simulation experiments (s)

| No. | No controls | PI-ALINEA | DRL | No. | No controls | PI-ALINEA | DRL |
| --- | --- | --- | --- | --- | --- | --- | --- |
| 1 | 186 | 179 [-4%] | 171 [-8%] | 11 | 188 | 173[-8%] | 167[-11%] |
| 2 | 203 | 173[-15%] | 171[-16%] | 12 | 211 | 182[-14%] | 170[-19%] |
| 3 | 217 | 175[-19%] | 177[-18%] | 13 | 219 | 175[-20%] | 172[-21%] |
| 4 | 227 | 181[-20%] | 181[-20%] | 14 | 170 | **172[+1%]** | 170[-0%] |
| 5 | 202 | 173[-14%] | 177[-12%] | 15 | 187 | 182[-3%] | 175[-6%] |
| 6 | 178 | **195 [+10%]** | 173[-3%] | 16 | 216 | 170[-21%] | 179[-17%] |
| 7 | 185 | 178[-4%] | 174[-6%] | 17 | 177 | **192[+8%]** | 174[-2%] |
| 8 | 226 | 180[-20%] | 173[-23%] | 18 | 228 | 169[-26%] | 172[-25%] |
| 9 | 175 | **214 [+22%]** | 169[-3%] | 19 | 190 | 179[-6%] | 175[-8%] |
| 10 | 206 | 183[-11%] | 175[-15%] | 20 | 194 | 180[-7%] | 170[-12%] |

Fig.9 presents the temporal-spatial distribution of the mean speeds along the mainline. The results further demonstrate the efficiency of the ramp metering and the better performance of the proposed method. In the no-control scenario, the minimum speed in the mainline is 12 m/s (43.2/h). The minimum speeds resulting from the PI-ALINEA method and the DRL method are 13.3% and 20% higher than that in the no-control scenario, respectively. The area with speed lower than 15 m/s (54km/h) is reduced with the PI-ALINEA method, which is further reduced when the DRL method is implemented.

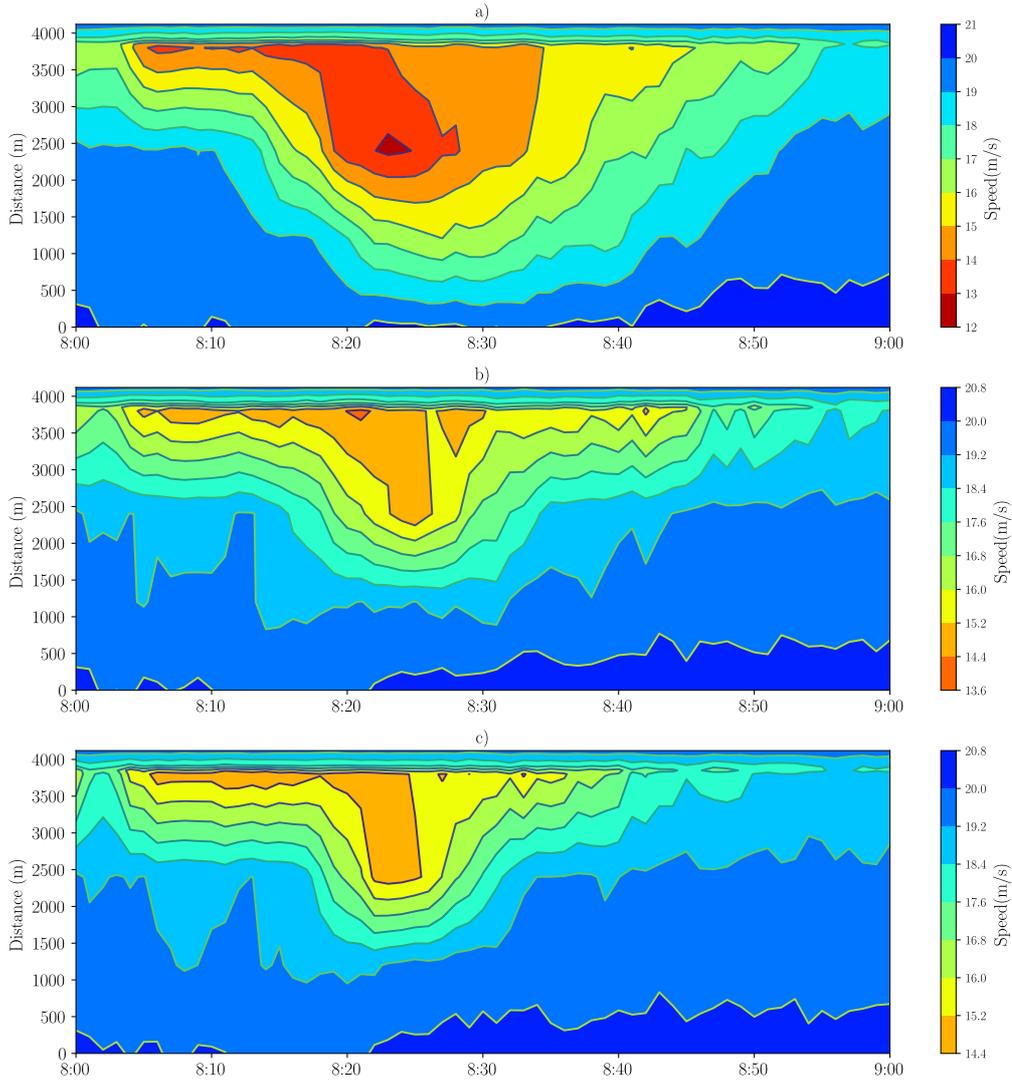

Figure 9: Temporal-spatial distribution of speed in a) No-control, b) PI-ALINEA c) DRL.

Fig. 10 presents the Q1, Q2, and Q3 of the queue lengths at the on-ramp in three scenarios. The results reveal that ramp metering improves vehicle mobility in the mainline at the cost of delays of vehicles at the on-ramp. Without ramp metering, no vehicle queues are formed at the on-ramp. Both ramp metering methods result in vehicle queues at the on-ramp. Nevertheless, the queues produced by the proposed method are shorter than those produced by the PI-ALINEA method. When the PI-ALINEA method and the proposed DRL method are adopted, the maximum values of the Q3 of the queue lengths are 126 m and 67 m, respectively.

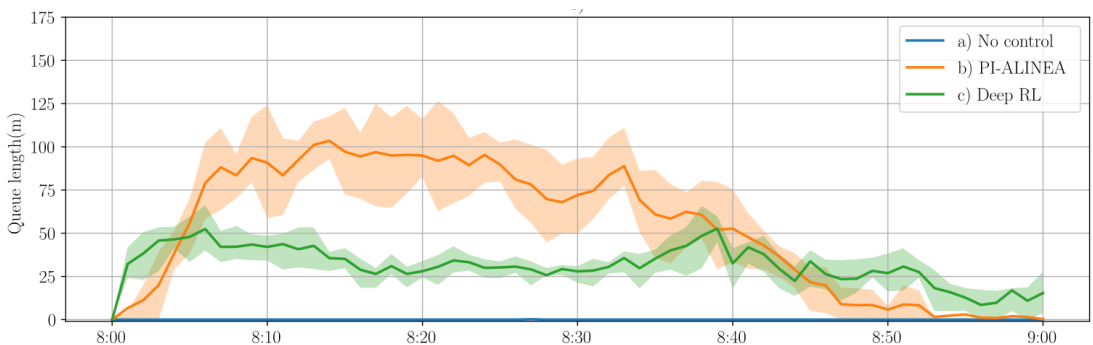

Figure 10: Queue Length at the on-ramp.

Fig.11 presents the queue lengths at the on-ramp and the occupancy downstream of the merging area for the two ramp metering methods. Fig.11a demonstrates that the queue length and the occupancy are highly correlated when the PI-ALINEA method is adopted. The results are understandable. Given high occupancy, the PI-ALINEA method tends to adopt a long red phase, which would lead to long queues at the on-ramp. In contrast, Fig.11b shows that the correlation between the queue length and the occupancy is relatively low when the DRL method is implemented. The queue length over time is relatively stable, ranging from 25-50 m. Although the occupancy is low after 8:45, the ramp meter is still active. The results indicate that the proposed DRL method not only relies on downstream occupancy but other information extracted from raw data.

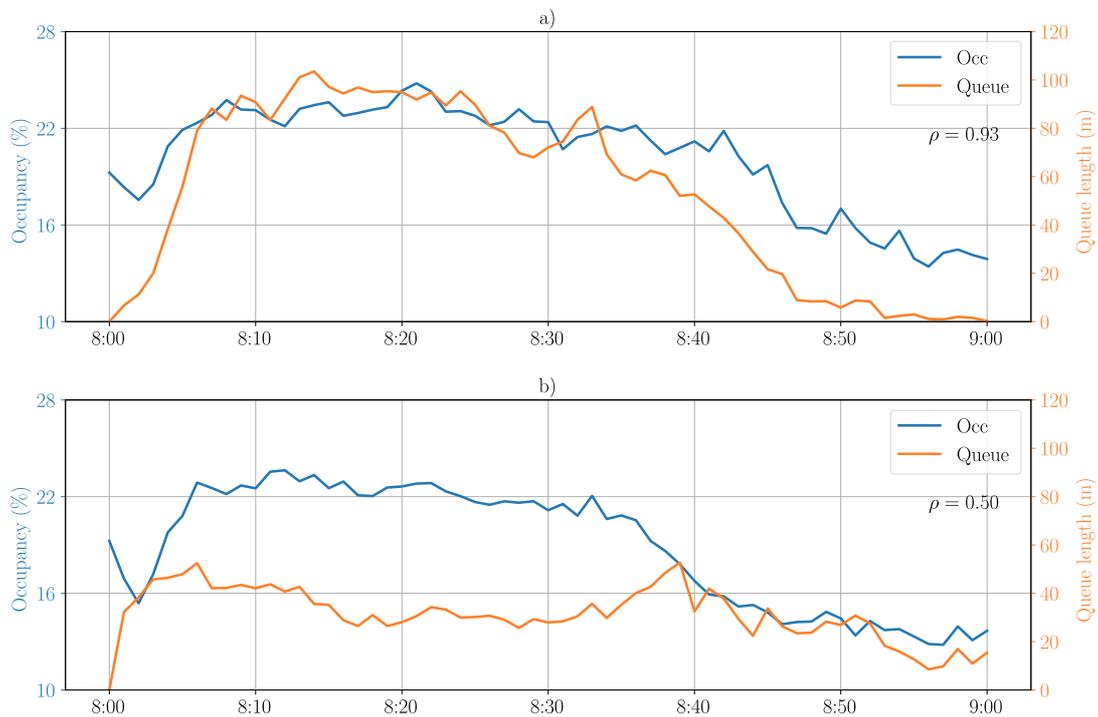

Figure 11: Queue length and occupancy for a) PI-ALINEA and b) DRL.

Fig.12 presents the mean queue lengths at the on-ramp and the red ratios (*i.e.*, the proportion of the red phase in the simulation time) in each simulation for the two ramp metering methods. The queue length generally increases with the red ratio. However, compared with the PI-ALINEA method, the queues produced by the DRL method are shorter, suggesting that the proposed method could better utilize the green time.

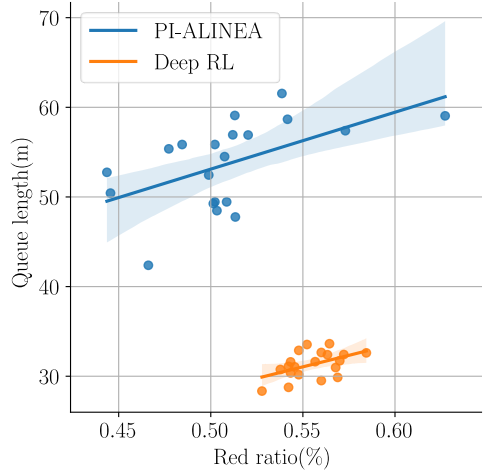

Figure 12: Relationship of queue length and red ratio for a) PI-ALINEA and b) DRL.

The better performance of the proposed DRL method is further revealed when considering traffic flows downstream of the merge area, which is demonstrated in Fig. 13. At the beginning traffic flows in three scenarios are close. During the peak period between 8:10-8:35, the flows resulting from the DRL method are higher and more stable, suggesting that the proposed DRL method could better alleviate the negative impacts of "capacity drop" resulting from massive merging behaviors. When traffic demand decreases during 8:40-9:00, traffic flows in the two ramp metering scenarios also decline. Nevertheless, traffic flows in the no-control scenario are still high during 8:45-8:55 since more vehicles are held in the system in the previous periods.

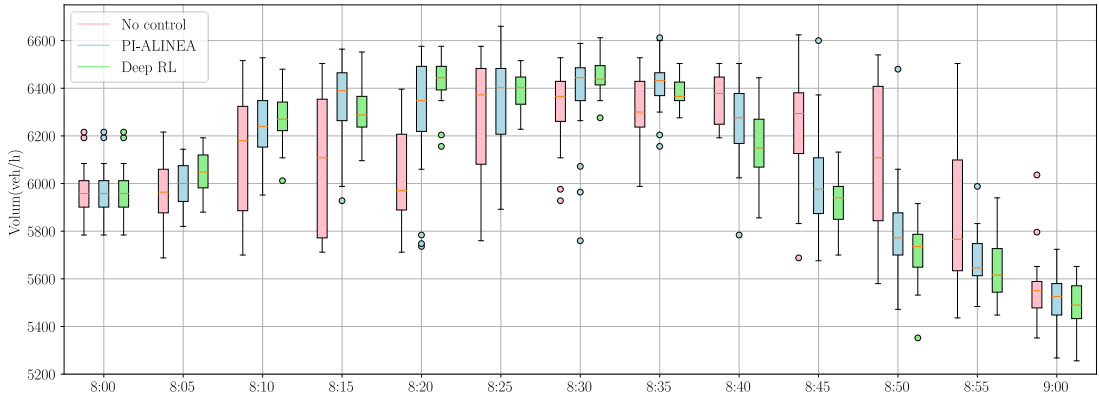

Figure 13: Downstream volume in three scenarios.

## Conclusion and Future Research

This study proposes a DRL method for local ramp metering based on traffic video data. The proposed method learns optimal strategies directly from high-dimensional visual inputs, overcoming the reliance on hand-crafted traffic measures. The better performance of the proposed method is demonstrated in a real-world case study. Compared with the traditional PI-ALINEA method, the proposed method results in lower travel times in the mainline, shorter vehicle queue lengths at the on-ramp, and could better alleviate the negative impacts of "capacity drop" resulting from massive merging behaviors. The results suggest that the proposed DRL method is able to extract useful information from the video data for better ramp metering controls.

Although promising, additional research would be required before the proposed method can be implemented for operational use. We assume that vehicle location information from the videos is accurate. Actually, the data quality may be disturbed by factors such as weather, vehicle sizes, and colors. It is valuable to design a ramp metering algorithm that is robust to those disturbances. In addition, abnormal events, such as accidents, are not considered in this study. Efficiently training the ramp metering to adapt to abnormal events remains to be investigated. Finally, extending the proposed method for coordinated ramp metering controls is worth pursuing.

## Data Availability

The data used to support the results of this study is available from the corresponding author upon request.

## Conflicts of Interest

The authors declare that there is no conflict of interest regarding the publication of this paper exists.

## Funding Statement

This work is supported by the National Natural Science Foundation of China (Grant Nos. 71671124, 71901164).